\documentclass[10pt,twocolumn,letterpaper]{article}

\usepackage{cvpr}
\usepackage{times}
\usepackage{epsfig}
\usepackage{graphicx}
\usepackage{amsmath}
\usepackage{amssymb}
\usepackage{bm}
\usepackage{color}
\usepackage{pbox}
\usepackage{array}

\usepackage[pagebackref=true,breaklinks=true,letterpaper=true,colorlinks,bookmarks=false]{hyperref}

\cvprfinalcopy 

\def\cvprPaperID{2466} 

\definecolor{ForestGreen}{RGB}{40,166,40}

\definecolor{darkgreen}{rgb}{0.0, 0.5, 0.0}

\def\eg{\emph{e.g.}\xspace}
\def\etal{\emph{et al.}\xspace}

\def\1{\mathds{1}}

\ifcvprfinal\fi
\begin{document}

\title{End-to-End Saliency Mapping via Probability Distribution Prediction}

\author{Saumya Jetley\\
University of Oxford\\
Oxford, United Kingdom\\
{\tt\small sjetley@robots.ox.ac.uk}
\and
Naila Murray\\
XRCE\\
Meylan, France\\
{\tt\small naila.murray@xrce.xerox.com}
\and
Eleonora Vig\\
XRCE\thanks{EV is now at the German Aerospace Center.}\\
Meylan, France\\
{\tt\small eleonora.vig@dlr.de}
}
\maketitle

\begin{abstract}
Most saliency estimation methods aim to explicitly model low-level conspicuity cues such as edges or blobs and may additionally incorporate top-down cues using face or text detection.
Data-driven methods for training saliency models using eye-fixation data are increasingly popular, particularly with the introduction of large-scale datasets and deep architectures.
However, current methods in this latter paradigm use loss functions designed for classification or regression tasks whereas saliency estimation is evaluated on topographical maps.
In this work, we introduce a new saliency map model which formulates a map as a generalized Bernoulli distribution.
We then train a deep architecture to predict such maps using novel loss functions which pair the softmax activation function with measures designed to compute distances between probability distributions.
We show in extensive experiments the effectiveness of such loss functions over standard ones on four public benchmark datasets, and demonstrate improved performance over state-of-the-art saliency methods.
\end{abstract}


\section{Introduction}\label{sec:intro}

This work is concerned with visual attention prediction, specifically, predicting a topographical visual saliency map when given an input image.
Visual attention has been traditionally used in computer vision as a pre-processing step in order to focus subsequent processing on regions of interest in images, an ever-more important step as vision models and datasets increase in size.
Saliency map prediction has found useful applications in tasks such as automatic image cropping \cite{stentiford2007attention}, content aware image resizing \cite{achanta2009saliency}, image thumb-nailing \cite{marchesotti2009framework}, object recognition \cite{gilanipet}, and fine-grained scene, and human action classification \cite{sharma2012discriminative}. 
Traditional saliency models, such as the seminal work of Itti \etal \cite{itti1998model}, have focused on designing mechanisms to explicitly model biological systems.
Another popular attention modelling paradigm involves using data-driven approaches to learn patch-level classifiers which give a local image patch a ``saliency score" \cite{KiWiScFr07, judd2009learning}, using eye-fixation data to derive training labels.
A recent trend has emerged which intersects with both of these paradigms: to use hierarchical models to extract saliency maps, with model weights being learned in a supervised manner.
In particular, end-to-end or ``deep" architectures, which have been successfully used in semantic labelling tasks such as categorization or object localization, have been re-purposed as attention models \cite{kummerer2014, junting15}.
This trend has been facilitated by the introduction of large visual attention datasets created using novel eye movement collection paradigms \cite{jiang2015salicon,xu15}.
However, while these deep methods have focused on designing appropriate architectures for extracting saliency maps, they continue to use loss functions adapted for semantic tasks, such as classification or regression losses.

\begin{figure}[!t]
\centering
\includegraphics[width=\linewidth]{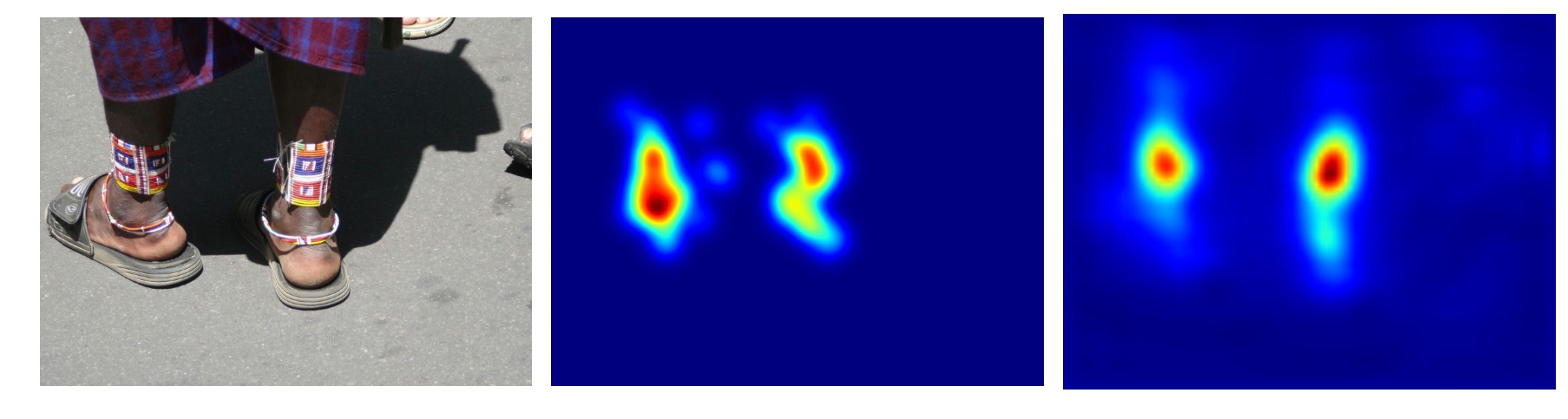}
\caption{Sample image (left) with ground-truth saliency map (middle) and map predicted by our PDP approach (right).}
\label{fig:teaser}
\end{figure}

In this work, we propose a novel formulation of saliency map prediction as a {\em probability distribution prediction} task.
The map is formulated as a generalized Bernoulli distribution, and several novel loss functions are proposed based on probability distance measures.
We show that training a deep architecture with such loss functions results in superior performance with respect to standard regression loss functions such as the Euclidean and Huber loss.
We also perform a comparison among our proposed loss functions and show that our loss function, based on the Bhattacharyya distance for multinomial distributions, gives top performance.

Our contributions are therefore the following:
\begin{itemize}
\item a novel formulation which represents a saliency map as a generalized Bernoulli distribution;
\item a set of novel loss functions which are paired with the softmax function and which penalize the distance between predicted and target distributions;
\item a fully-convolutional architecture which can generate a saliency map for a large image in 200ms using modern GPUs.
\end{itemize}
Our extensive experimental validation on four datasets demonstrates the effectiveness of our approach when compared to other loss functions and other state-of-the-art approaches to saliency map generation.
Figure~\ref{fig:teaser} illustrates its prediction performance.

The remainder of the paper is organized as follows: in section~\ref{sec:related} we discuss related work.
Section~\ref{sec:method} describes our saliency modelling and estimation approach. We report and discuss evaluation results in section~\ref{sec:results} and conclude in section~\ref{sec:conclusion}.


\section{Related work}\label{sec:related}

Existing approaches can be organized into one of four 
broad categories based on whether they involve a \textit{shallow} or \textit{deep}
architecture, and an \textit{unsupervised} or \textit{supervised} learning
paradigm. We will discuss each of these broad categories in turn.
For an excellent survey of saliency estimation methods, please refer to \cite{borji2013state}.\\[2mm]
\textbf{Unsupervised shallow methods}
Most early work on saliency builds on psychological and
psychophysical models of attention as studied in humans. Koch and Ullman
\cite{koch1987shifts} were among the first to use feature integration
theory \cite{treisman1980feature} to propose a set of individual
topographical maps of elementary cues such as color, contrast, and
motion, and combine them to produce a global topographical map of saliency. Their model is
implemented using a simple neural circuitry with winner-take-all and
inhibition-of-return mechanisms. It is further investigated in \cite{itti2000saliency} by combining features maps over a wider set of modalities (42 such maps) and testing on real-world images. Later approaches largely explore the same idea of complementary feature ensembles 
\cite{itti1998model,isocentriccolor&curvedness,liu2013saliency_regionalhist,depthmatters,zhang2013saliency,murray2013low}
and often add to it additional center-surround cues~\cite{itti1998model,murray2011saliency,zhang2013saliency}. 

Complementing the biologically motivated approaches, a number of methods
adopt an information-theoretic justification for attentional selection, \eg
by self-informa\-tion~\cite{ZhToMaShCo08}, information
maximization~\cite{BrTs06}, or Bayesian surprise~\cite{ItBa06}.
High computational efficiency is achieved by spectrum-based methods~\cite{HoZh07,ScSt12}.
All these approaches use bottom-up cues, are shallow (one or few layers) and involve no or minimalistic learning of
thresholds/heuristics.\\[2mm]
\textbf{Supervised shallow methods}
This category includes learning based approaches involving models such as
markov chains \cite{harel2006graph}, support vector machines
\cite{KiWiScFr07,judd2009learning} and adaboost classifiers \cite{cerf2008predicting}.
\cite{harel2006graph} substitutes the idea of centre-surroundedness and
normalization with learnable graph weights. \cite{cerf2008predicting}, 
\cite{judd2009learning}, and \cite{ZhKo11} enrich learning by incorporating
top-down semantic cues in the form of detection maps for faces, persons,
cars, and the horizon.\\[2mm]
\textbf{Unsupervised hierarchical methods}
In the context of saliency prediction,
the first attempts to employ deeper architectures are mostly unsupervised.
\cite{ShSoZh12} learn higher-level concepts from fixated image patches
using a 3-layer network of sparse coding units.  
\cite{vig2014large} perform a large-scale search for optimal
network architectures of up to three layers, but the network weights are
not learned. 

DeepGaze~\cite{kummerer2014} employs an existing network
architecture, the 5-layer deep AlexNet~\cite{KrSuHi12} trained for object
classification on ImageNet, to demonstrate that off-the-shelf CNN features
can significantly outperform non-deep and ``shallower'' models, even if not
trained explicitly on the task of saliency prediction. Learning, in their
case, has meant finding the optimal linear combination of features
from the different network layers.\\[2mm]

\textbf{Supervised hierarchical methods}
The publication of large-scale attention datasets, such as
SALICON~\cite{jiang2015salicon} and TurkerGaze/iSUN~\cite{xu15}, has enabled training deep architectures specifically for the task of saliency prediction. Our work lies in this category and involves training an end-to-end deep model with a novel loss function.

SALICON~\cite{jiang2015salicon} was collected with a new data-collection
paradigm, in which observers were shown foveated images and were asked to
move the mouse cursor around to simulate the high-resolution fovea.
This novel paradigm was used to annotate 20K images from the MSCOCO dataset~\cite{Lin2014MScoco}. Relying on this new large-scale dataset, the authors of \cite{junting15} trained a network end-to-end for saliency prediction. Their network, titled JuntingNet, consists of five convolutional and two fully-connected layers, and the parameters of the network are learned by
minimizing the Euclidean loss function defined on the ground-truth saliency
maps. This method reports state-of-the-art results on the LSUN 2015
saliency prediction challenge~\cite{salicon-saliency-benchmark}. 

Another end-to-end approach that formulates saliency prediction as regression is
that of~\cite{deepfix15}. DeepFix builds upon the very deep
\mbox{VGGNet}~\cite{Simonyan14c}, uses
convolutional layers with large and multi-size receptive fields to capture
complementary image context, and introduces a location-biased convolutional (LBC) layer to model the center-bias. 

Finally, one of the most recent works in this paradigm \cite{huang2015salicon} proposes the use of deep neural networks to bridge the semantic gap in saliency prediction via a two-pronged strategy. The first is the use of the KL-divergence as a loss function motivated by the fact that it is a standard metric for evaluation of saliency methods. The second is the aggregation of response maps from both coarse and fine resolutions. 
In this work, we argue for a well-motivated probabilistic modelling of the saliency maps and hence study the use of KL-divergence, among other probability distance measures, as loss functions. As we discuss in section~\ref{sec:results}, we observe that our Bhattacharyya distance-based loss function consistently outperforms the KL-divergence-based one across 4 standard saliency metrics.


\section{Saliency maps as probability distributions}\label{sec:method}

\begin{table*}[!ht]
\centering
\begin{tabular}{|l||c|c|}
\hline
\textbf{Probability distances} & $L(\bm{p},\bm{g})$ & $\frac{\partial L(\bm{p},\bm{g})}{\partial x^p_{i}}$ \\
\hline
\hline
$\chi^2$ divergence & $\sum_j\frac{(g_j)^2}{p_j}-1$ & $p_i\sum_{j\neq i}\frac{g^2_j}{p_j} - \frac{g^2_i}{p_i}(1-p_i)$ \\
\hline
Total Variation distance & $\frac{1}{2}\sum_j|g_j - p_j|$ & $\frac{1}{2}\left[ p_i\sum_{j\neq i}\frac{g_j - p_j}{|g_j - p_j|}p_j - p_i\frac{g_i - p_i}{|g_i - p_i|}(1 - p_i) \right]$ \\
\hline
Cosine distance & $1 - \frac{\sum_jp_jg_j}{\sqrt{\sum_jp^2_j}\sqrt{\sum_jg^2_j}}$ & \pbox{.6\textwidth}{$\frac{1}{C}\left[ p_i\sum_{j\neq i}p_j(g_j - p_i\frac{\sqrt{\sum_ig^2_i}}{\sqrt{\sum_ip^2_i}}R) - p_i(g_i - p_iR)(1-p_i) \right]$; \\ where $R=\frac{\sum_ip_ig_i}{C}$ and $C=\sqrt{\sum_ip^2_i}\sqrt{\sum_ig^2_i}$.} \\
\hline
Bhattacharyya distance & $-\ln\sum_j(p_jg_j)^{0.5}$ & $\frac{-1}{2\sum_j (p_jg_j)^{0.5}}\left[ p_i\sum_{j\neq i}(p_jg_j)^{0.5} - (p_ig_i)^{0.5}(1-p_i) \right]$ \\
\hline
\hline
KL divergence & $\sum_j g_j\log\frac{g_j}{p_j}$ & $p_i\sum_{j\neq i}g_j - g_i(1-p_i)$ \\
\hline
\end{tabular}
\vspace{5pt}
\caption{Probability distance measures and their derivatives used for stochastic gradient descent with back-propagation. We propose the use of the first 4 meausres as loss functions. We also investigate KL-divergence, which is widely used to train recognition models in the form of the closely-related cross-entropy loss.}
\label{tab:losses}
\end{table*}

Saliency estimation methods have typically sought to model local saliency based on conspicuity cues such as local edges or blob-like structures, or on the scores of binary saliency classifiers trained on fixated and non-fixated image patches.
More recently, methods have sought to directly predict maps using pixel-wise regression.

However, visual attention is a fundamentally stochastic process due to it being a perceptual and therefore subjective phenomenon.
In an analysis of 300 images viewed by 39 observers, the authors of \cite{Judd_2012} find that the fixations for a set of $n$ observers match those from a different set of $n$ observers with an AUC score that increases with the increase in the value of $n$. The lower bound of human performance is found to be $85\%$ AUC. Therefore there is high consistency across observers.
At the limit of $n\rightarrow\infty$ this AUC score is $~92\%$, which can therefore be considered a realistic upper-bound for saliency estimation performance.

Ground-truth saliency maps are constructed from the aggregated fixations of multiple observers, ignoring any temporal fixation information.
Areas with a high fixation density are interpreted as receiving more attention.
As attention is thought to be given to a localized region rather than an exact pixel, two-dimensional Gaussian filtering is typically applied to a binary fixation map to construct a smooth ``attentional landscape" \cite{zangemeister1996visual} (\cf Figure~\ref{fig:teaser}, middle image for an example). Our goal is to predict this attentional landscape, or saliency map.
Given the stochastic nature of the fixations upon which the maps are based, and the fact that the maps are based on aggregated fixations without temporal information, we propose to model a saliency map as a probability distribution over pixels, where each value corresponds to the probability of that pixel being fixated upon. That is, we represent a saliency map as a generalized Bernoulli distribution $\pmb{p} = (p_1,\cdots,p_i,\cdots,p_N)$, where $\pmb{p}$ is the probability distribution over a set of pixels forming an image, $p_i$ is the probability of pixel $i$ being fixated upon and $N$ is the number of image pixels.
While this formulation is somewhat simplistic, it will allow for novel loss functions highly amenable to training deep models with back-propagation. In the sequel, we first describe these loss functions and then describe our model implementation.

\subsection{Learning to predict the probability of fixation}
We adopt an end-to-end learning framework in which a fully-convolutional network is trained on pairs of images and ground-truth saliency maps $\pmb{g}$ modeled as distributions.
The network outputs predicted distributions $\pmb{p}$\footnote{We slightly abuse notation and from now on use $\pmb{p}$ to refer specifically to the predicted distribution.}.
Both probability distributions, $\pmb{g}$ and $\pmb{p}$, are computed using the softmax activation function:
\begin{equation}
p_i = \frac{e^{x^p_i}}{\sum_j e^{x^p_j}},\qquad g_i = \frac{e^{x^g_i}}{\sum_j e^{x^g_j}},
\end{equation}
where $\pmb{x} = (x_1,\cdots,x_i,\cdots,x_N)$ is the set of un-normalized saliency response values for either the ground-truth map ($\pmb{x^g}$) or the predicted map ($\pmb{x^p}$).
To compute $\pmb{x^g}$, a binary fixation map $\pmb{b}$ is first generated from ground-truth eye-fixations.
The binary map $\pmb{b}$ is then convolved with a Gaussian kernel as described earlier in this section to produce $\pmb{y}$.
The smoothed map $y$ is then normalized as
\begin{equation}
{x^g_i} = \frac{y_i - \min[\bm{y}]}{\max[\bm{y}] - \min[\bm{y}]}.
\end{equation}

We generate $\pmb{x^p}$ directly from the last response map of our deep network, whose architecture is described in the next section.

We propose to combine the softmax function with distance measures appropriate for probability distributions in order to construct objective functions to be used for training the network.
This combination is inspired by the popular and effective softmax/cross-entropy loss pairing which is often used to train models for multinomial logistic regression.

In our case, we propose to combine the softmax functions with the $\chi^2$, total-variation, cosine and Bhattacharyya distance measures, as listed in Table~\ref{tab:losses}.
To our knowledge, these pairings have not previously been used to train a network for probability distribution prediction.
We also investigate the use of the KL divergence measure, the minimization of which is equivalent to cross-entropy minimization, and which is used extensively to learn regression models in deep networks.
The partial derivatives of these loss functions with respect to $x^p_i$ are all of the form $ap_i - b(1-p_i)$ due to the pairing with the softmax function, whose partial derivative with respect to $x^p_i$ is
\begin{equation}
\frac{\partial p_j}{\partial x^p_i}=
\begin{cases}
p_i(1 - p_i),& \text{if } j = i \\
-p_ip_j, & \text{otherwise}.
\end{cases}
\end{equation}

We make comparisons with two standard regression losses, the Euclidean and Huber losses, defined as:
\begin{equation}
L_{euc}(\bm{p},\bm{g})= \sum_j a_j^2,
\end{equation}
and
\begin{equation}
L_{hub}(\bm{p},\bm{g})= \sum_j
\begin{cases}
\frac{1}{2}a_j^2,& \text{for $|a_j| \leq 1$ } \\
|a_j| - \frac{1}{2}, & \text{otherwise};
\end{cases}
\end{equation}

where $a_j = |p_j-g_j|$.

\subsection{Training the prediction model}

\begin{figure*}[ht]
\centering
\includegraphics[width=1\textwidth]{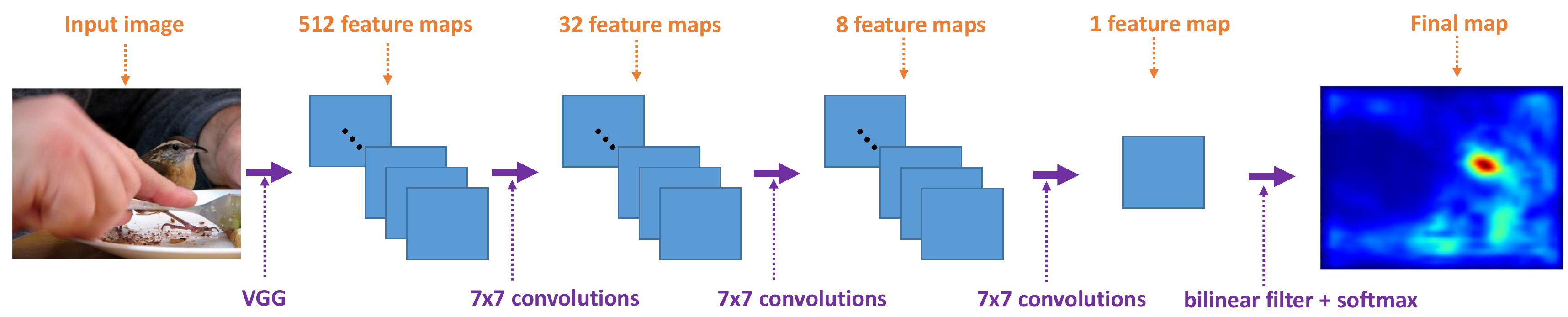}
\caption{Our proposed saliency map extraction pipeline: the input image is introduced into a convNet with an identical architecture to the convolutional-layer portion of VGGNet. Additional convolutional layers are then applied, resulting in a single response map which is upsampled and softmax-normalized at testing time to produce a final saliency map.}
\label{fig:pipeline}
\end{figure*}

The network architecture and saliency map extraction pipeline is shown in Figure~\ref{fig:pipeline}.
We use the convolutional layers of the VGGNet model~\cite{Simonyan14c}, which were trained on ImageNet images for the task of classification, as the early layers of our model.
This convolution sub-network has been shown to provide good local feature maps for a variety of different tasks including object localization 
\cite{DBLP:journals/corr/RenHG015} and semantic segmentation~\cite{Long_2015_CVPR}.
As saliency datasets tend to be much too small to train such large networks from random initializations (the largest dataset has 15000 images, compared to 1M for ImageNet), it is essential to initialize with a pre-trained network.
We then progressively decrease the number of feature maps using additional convolutional layers, until a final down-sampled saliency map is produced.
We add three new layers, rather than just one, to predict the final map in order to 
improve both discriminability and generalizability~\cite{Simonyan14c}. We experimented with different filter sizes besides $7\times7$ (\eg $9\times9$, $5\times5$, $3\times3$) and found no significant performance difference. We explicitly avoided fully-connected layers in order to obtain a memory and time-efficient model. 
The three new layers are initialised with a uniform Gaussian distribution of sigma = 0.01.
Because the response maps undergo several max-pooling operations, the predicted saliency map $\pmb{p}$ is lower-resolution than the input image. The ground-truth map $\pmb{g}$ is therefore downsampled during training to match the dimensions of $\pmb{p}$. Conversely, during inference the predicted map is upsampled with a bilinear filter to match the dimensions of the input image (see Figure~\ref{fig:pipeline}), and the softmax function is applied for normalization to a probability distribution.

The final fully-convolutional network comprises 16 convolutional layers, each of which is followed by a ReLu layer.
Due to the fully-convolutional architecture, the size is quite small for a deep model, with only 15,530,481 weights (60MB of disk space).

Note that while several deep saliency models explicitly include a center bias (see e.g. \cite{deepfix15}), we hypothesized that the model could learn the center-bias implicitly, given that it is largely an artifact of a composition bias in which photographers tend to place highly salient objects in the image center~\cite{borji2015}. We tested this by adding Gaussian blurring and a center-bias to our maps, with optimized parameters, using the post-processing code of the MIT saliency benchmark~\cite{mit-saliency-benchmark}. We found no consistent improvement across different metrics using this post-processing which indicates that a great deal of center-bias and Gaussian blurring is already accounted for in the model.

The objective function is optimized using stochastic gradient descent, with a learning rate of 1 times the global learning rate for newly-introduced layers and 0.1 times the global learning rate for those layers which have been pre-trained on ImageNet.
To reduce training time, the first 4 convolutional layers were fixed and thus retained their pre-trained values.
We used a momentum of 0.9 and a weight decay of 0.0005. The model is implemented in Caffe~\cite{jia2014caffe}.
We trained the network using an Nvidia K40 GPU.
Training on the SALICON training set took 30 hours.

Saliency datasets tend to have semantic biases and other idiosyncrasies related to the complexity of collecting eye-tracking information (such as the viewing distance to the screen and the eye-tracker calibration).
For this reason, we perform dataset-specific fine-tuning, which improves performance.
Fine-tuning is particularly essential in our case because the SALICON dataset collected mouse clicks in lieu of actual eye-fixations which, while highly correlated in general, are still an approximation to true human eye movements. As shown on a subset of the SALICON images, image-level conformance between SALICON fixations and human eye fixations can be as low as shuffled AUC (sAUC) of 0.655 and as high as sAUC of 0.965 \cite{jiang2015salicon}.
Therefore it is beneficial to fine-tune the network for each dataset of interest.
A detailed description of each of these datasets follows.


\section{Experimental evaluation}\label{sec:results}
This section describes the experimental datasets used for training and evaluating the saliency prediction models followed by a discussion on the quantitative and qualitative aspects of the results.

\subsection{Datasets}

\paragraph{SALICON}
This is one of the largest saliency datasets available in the public domain~\cite{jiang2015salicon}. It consists of eye-fixation information for 20000 images from the MS COCO dataset~\cite{Lin2014MScoco}. These images contain diverse indoor and outdoor scenes and display a range of scene clutter. 10000 images are marked for training, 5000 for validation and 5000 for testing. The fixation data for the test set is held-out and performance on it must be evaluated on a remote server. 
The peculiarity of SALICON lies in its mouse-based paradigm for fixation gathering. The attentional focus (foveation) in the human attention mechanism that defines saliency fixations is simulated using mouse-movements over a blurred image. The approximate foveal image region around the mouse position is selectively un-blurred as the user explores the image scene using the mouse cursor. As evaluated on a subset of the dataset, this mouse-click data is in general highly consistent with human eye fixations (at 0.89 sAUC). Therefore, while the mouse fixation data is an approximation to the human baseline, it is useful in adapting the weights of a deep network originally trained for a distinct task to the new task of saliency prediction.
We use this dataset for our comparative study of the selected probability distances as loss functions during learning. We have also submitted our best performing model to the SALICON challenge server~\cite{salicon-saliency-benchmark}.\\[2mm]
\noindent\textbf{MIT-1003}
This dataset was introduced as part of the training and testing paradigm in \cite{judd2009learning}. The eye tracking data is collected using a head-mounted eye tracking device for 15 different viewers. The 1003 images of this dataset cover natural indoor and outdoor scenes. For our experiments, we use the first 900 images for training and the remaining 103 for validation, similar to the paradigm of \cite{deepfix15}.\\[2mm]
\noindent\textbf{MIT-300}
This benchmark consists of held-out eye tracking data for 300 images collected across 39 different viewers~\cite{Judd_2012}. The data collection paradigm for this dataset is very similar to that used in MIT-1003. Hence, as suggested on the online benchmark, we use MIT-1003 as the training data to fine-tune for MIT-300.\\[2mm]
\noindent\textbf{OSIE}
This benchmark contains a set of 700 images. These include natural indoor and outdoor scenes, as well as high aesthetic-quality pictures taken from Flickr and Google. In order to gain from top-down understanding, this dataset provides object and semantic level information (which we do not use) along with the eye-tracking data. Following the work of~\cite{Luo_2015_CVPR}, we randomly divide the set into 500 training and 200 test images and average the results over a 10-fold cross-validation.\\[2mm]
\noindent\textbf{VOCA-2012}
With the exception of SALICON, the previous datasets are relatively small, with at most 1003 images. Evaluations on large-scale datasets of real fixations would be more informative. However, to our knowledge, there is no truly large-scale dataset of free-viewing fixations. 
Instead, we evaluate on VOCA-2012, an action recognition dataset which has been augmented with task-dependent eye-fixation data~\cite{Mathe13}.
Predicting such fixations is a different task to predicting free-viewing fixations, the task for which our model is designed.
We therefore evaluate on this dataset to determine whether our model generalizes to this task.\\[2mm]
\noindent\textbf{Generating ground-truth maps}
To create ground-truth saliency maps from fixation data, we use the saliency map generation parameters established by the authors of each dataset. For SALICON, this means convolving the binary fixation maps with a Gaussian kernel of width 153 and standard deviation 19. For OSIE, this means applying a Gaussian kernel of width of 168 and standard deviation of 24 (all in units of pixels). The authors of MIT-1003 and MIT-300 provide ground-truth saliency maps which, according to their technical report~\cite{Judd_2012}, are computed with a Gaussian kernel whose size corresponds to a cutoff frequency of 8 cycles per image.

\subsection{Results}

We first compare results for different loss functions and then compare to the state-of-the-art methods.
For each dataset, we follow the established evaluation protocol and report results on standard saliency metrics, including sAUC, AUC-Judd, AUC-Borji, Correlation Coefficient (CC), Normalized Scanpath Saliency (NSS), Similarity (SIM), and Earth Mover's Distance (EMD).

\begin{figure}[ht]
\centering
\includegraphics[width=0.495\linewidth]{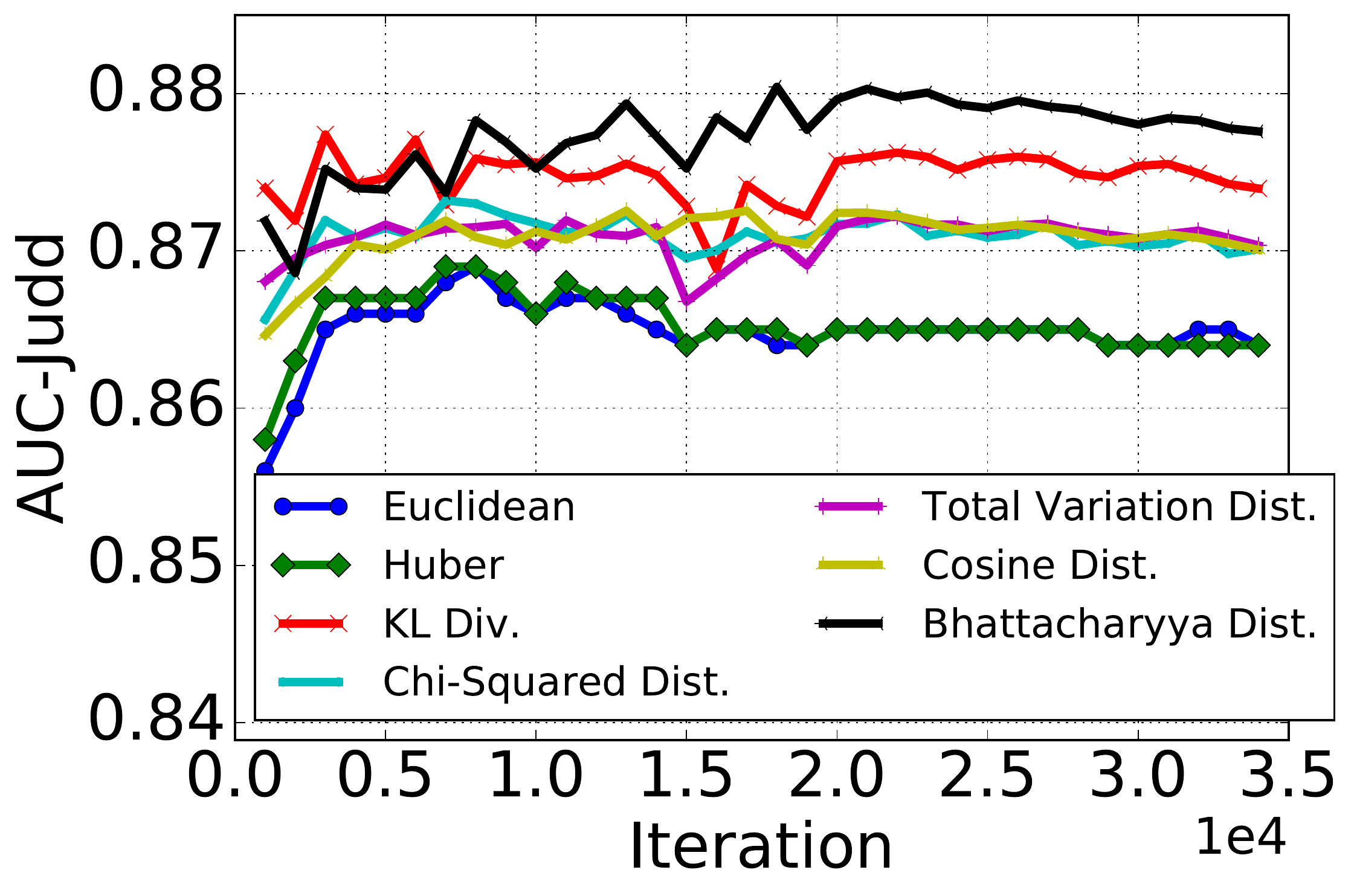}
\includegraphics[width=0.495\linewidth]{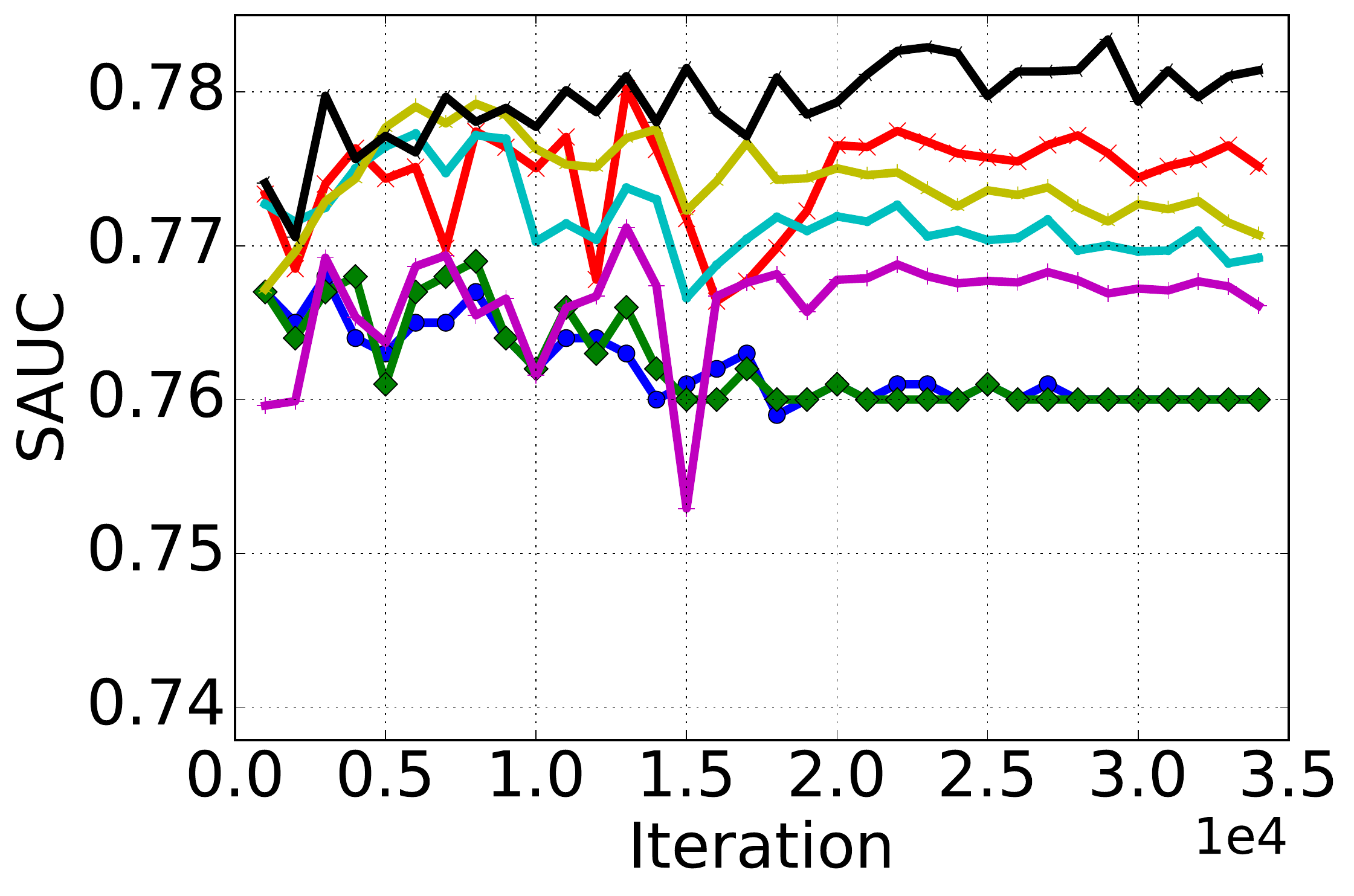}
\includegraphics[width=0.495\linewidth]{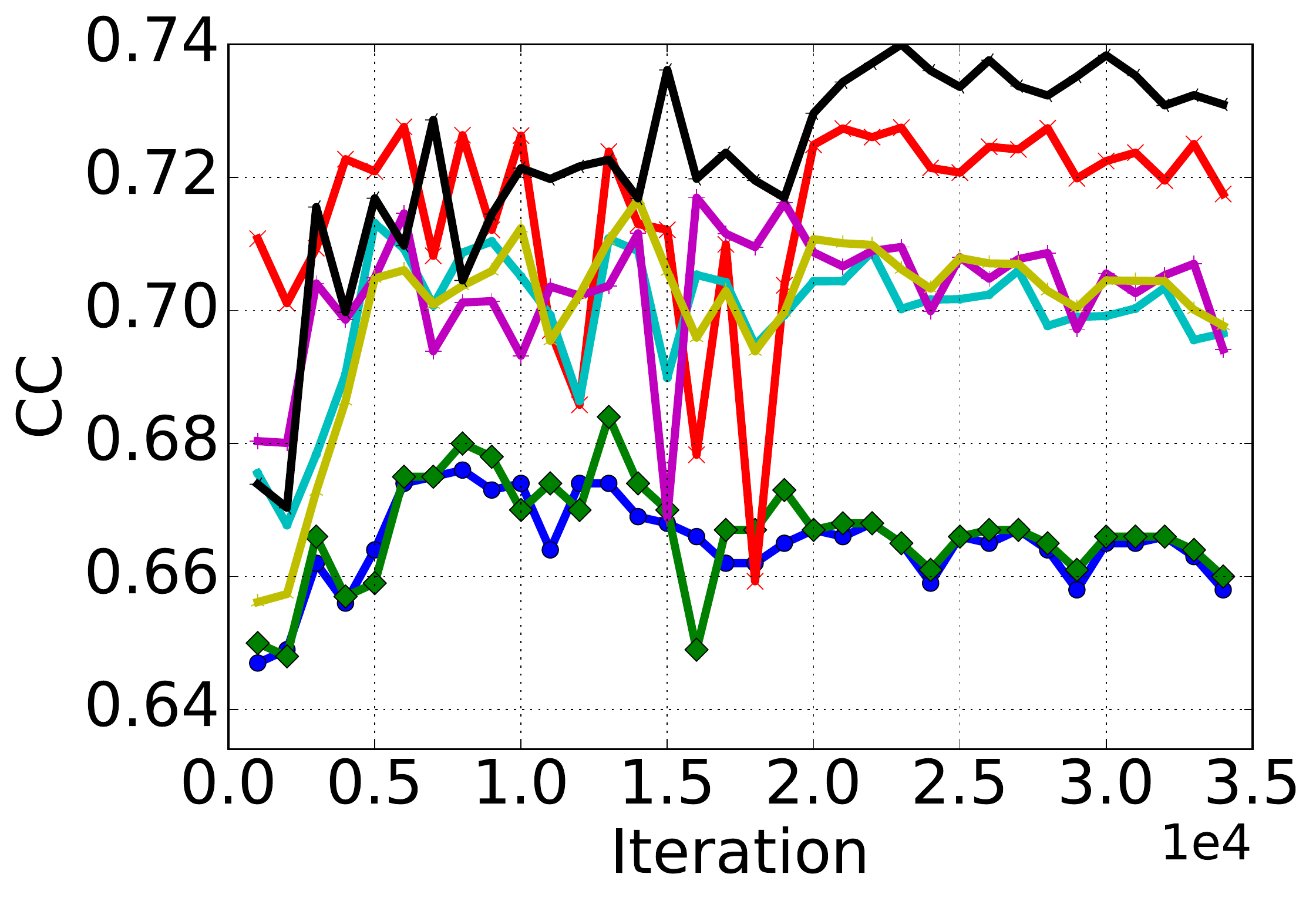}
\includegraphics[width=0.495\linewidth]{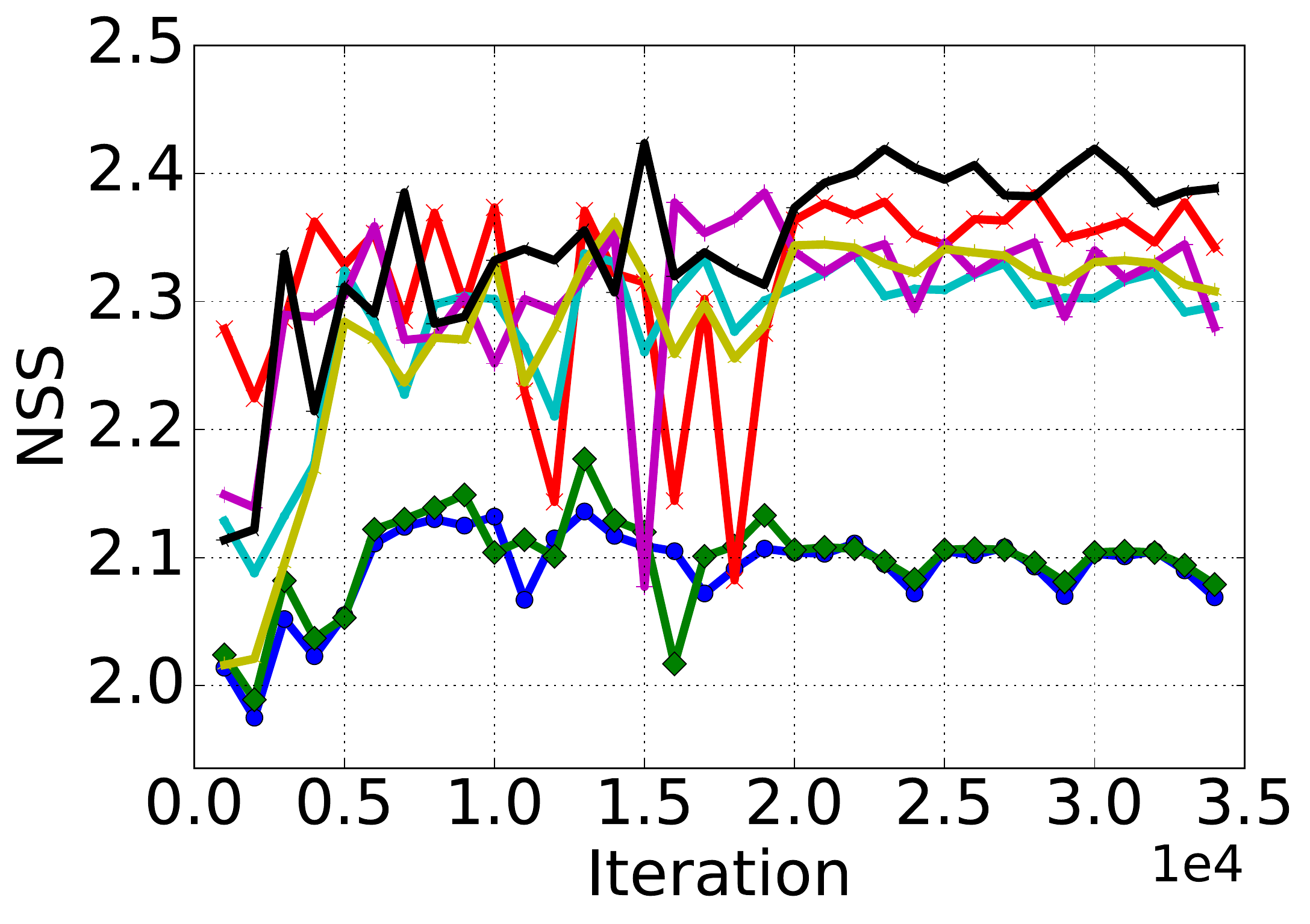}
\caption{Evolution of different metrics on the evaluation set of SALICON as the number of training iterations increases.}
\label{fig:convergence}
\end{figure}

\paragraph{Loss functions}
We compare the performance of models trained using our proposed loss functions to those trained on standard loss functions based on the Euclidean distance, Huber distance, and KL-divergence measure.
These models are all trained on the SALICON training set of 10K images, and validated on the SALICON validation set of 5K images.
Table~\ref{tab:sal_val} presents the best validation-set performance for each loss, as measured by the overall performance with respect to 4 metrics. These results show that: (i) the losses based on distance measures appropriate for probability distributions perform better than standard regression losses; (ii) the KL-divergence compares favorably with other methods; and (iii) the Bhattacharyya distance-based loss outperforms all other losses.
These two last losses share the property that they are robust to outliers as they suppress large differences between probabilities (logarithmically in the case of the KL divergence and geometrically in the case of the Bhattacharyya distance).
This robustness is particularly important as the ground-truth saliency maps are derived from eye-fixations which have a natural variation due to the subjectivity of visual attention, and which may also contain stray fixations and other noise.
Figure~\ref{fig:convergence} shows the evolution of the saliency metrics on the SALICON validation set as the training progresses. The Bhattacharyya distance is consistently the best-performing.

\begin{table}
\footnotesize
\centering
\renewcommand{\arraystretch}{1.2}
\begin{tabular}{|c|c|c|c|c|}
\hline
{\bf Distance} & AUC-Judd & sAUC & CC & NSS \\
\hline
\hline
Euclidean & 0.865 & 0.761 & 0.667 & 2.108\\
Huber & 0.867 & 0.766 & 0.684 & 2.177\\
KL divergence & 0.876 & 0.780 & 0.724 & 2.371\\
$\chi^2$ divergence & 0.872 & 0.774 & 0.711 & 2.337\\
Total Variation distance & 0.869 & 0.766 & 0.716 & 2.385\\
Cosine distance & 0.871 & 0.778 & 0.717 & 2.363\\
Bhattacharyya distance & {\bf 0.880} & {\bf 0.783} & {\bf 0.740} & {\bf 2.419}\\
\hline
\end{tabular}
\vspace{2mm}
\caption{SALICON validation set: Performance comparison of models trained using different loss functions.}
\label{tab:sal_val}
\end{table}

\paragraph{Comparison to the state of the art}
We compare the performance of our proposed model, using the Bhattacharyya distance, with the state-of-the-art methods for four standard saliency benchmarks as follows.

\emph{SALICON challenge:}
The saliency estimation challenge~\cite{salicon-saliency-benchmark} consists in predicting saliency maps for 5000 images held out from the SALICON dataset.
Table~\ref{tab:saltest} shows results for state-of-the-art methods and our approach, which we call PDP for probability distribution prediction.
We outperform all published results, to our knowledge, on this dataset across all three metrics.

\begin{table}
\footnotesize
\setlength\tabcolsep{6pt}
\renewcommand{\arraystretch}{1.2}
\centering
\begin{tabular}{|l|c|c|c|c|c|}
\hline
Method 	& CC 	& sAUC &	AUC-Borji \\
\hline
\hline
Itti\cite{itti1998model} &	0.205 &	0.610 &	0.660\\
GBVS\cite{harel2006graph} &	0.421 &	0.630 &	0.782 \\
BMS\cite{zhang2013saliency} &	0.427 &	0.694 &	0.770  \\
WHU\_IIP* &	0.457 &	0.606 &	0.776 \\
Xidian* &	0.481 &	0.681 &	0.800 \\
Rare12\_Improved* &	0.511 &	0.664 &	0.805 \\
UPC\cite{junting15} &	0.596 &	0.670 &	0.829 \\
\hline
PDP & {\bf 0.765} & {\bf 0.781} & {\bf 0.882} \\
\hline 
\end{tabular}
\vspace{2mm}
\caption{SALICON Challenge: comparison between different methods. Methods marked by * have no associated publication to-date.}
\label{tab:saltest}
\end{table}

\emph{MIT-300:}
MIT-1003 images serve as the training set for fine-tuning to this benchmark. The results are compared in Table~\ref{tab:mit300}. We perform comparably to the state-of-the-art methods. Note that DeepFix~\cite{deepfix15} incorporates external cues such as center and horizon biases in its models.
We believe that including such cues may also improve our model.
In addition, they use a larger architecture, but train with a regression loss.
Therefore our approach may complement theirs.
Fine-tuning on MIT-1003 could only be performed using a batch size of 1 image due to the large variations in size and aspect ratio of the images. We observed that a much-reduced momentum of 0.70 improved stability and allowed for an effective learning of the model with this constraint.

\begin{table}
\footnotesize
\setlength\tabcolsep{2pt}
\renewcommand{\arraystretch}{1.2}
\begin{tabular}{|l|c|c|c|c|c|c|c|}
\hline 
Method & AUC-Judd & SIM & EMD & AUC-Borji & sAUC & CC & NSS \\
\hline
\hline
eDN\cite{vig2014large} & 0.82 & 	0.41 &	4.56 & 	0.81 &	0.62 	& 0.45 	& 1.14 \\
BMS\cite{zhang2013saliency} & 0.83 	& 0.51 & 	3.35 &	0.82 &	0.65 	& 0.55 & 	1.41 \\
SALICON\cite{huang2015salicon} & {\bf 0.87} &	0.60 	& 2.62 &	0.85 &	{\bf 0.74} 	& 0.74 & 	2.12 \\
DeepFix\cite{deepfix15} & 0.{\bf 87} 	& {\bf 0.67} &	{\bf 2.04} &	{\bf 0.80} &	0.71 &	{\bf 0.78} &	{\bf 2.26} \\ 
\hline
PDP & 0.85 & 0.60 & 2.58 & 0.80 & 0.73 & 0.70 & 2.05 \\ 
\hline 
\end{tabular}
\vspace{2mm}
\caption{MIT-300: comparison with the state of the art.}
\label{tab:mit300}
\end{table}

\emph{OSIE benchmark:}
The performance comparison on this dataset is done using 10-fold cross validation by randomly dividing the dataset into 500 training and 200 validation images. Table~\ref{tab:osie700} shows that PDP achieves the highest sAUC score. This dataset contains a wide variety of image content and aesthetic properties. Nonetheless, this small set of 500 images was sufficient to successfully adapt our model.

\begin{table}
\footnotesize
\centering
\setlength\tabcolsep{3pt}
\renewcommand{\arraystretch}{1.2}
\begin{tabular}{|l|c|}
\hline 
Method & sAUC \\
\hline
\hline
Itti\cite{itti1998model} & 0.658 \\
SUN\cite{ZhToMaShCo08} & 0.735 \\
Signature\cite{hou2012image} & 0.749 \\
GBVS\cite{harel2006graph} & 0.706 \\
LCQS-baseline\cite{Luo_2015_CVPR} & 0.765 \\
\hline
PDP & {\bf 0.797} \\ 
\hline 
\end{tabular}
\vspace{2mm}
\caption{OSIE: The performance metric of shuffled AUC (sAUC) is averaged over 10-fold cross validation. (Baseline results are taken from~\cite{Luo_2015_CVPR}.)}
\label{tab:osie700}
\end{table}

\emph{VOCA-2012 (Generalization to task-dependent fixation prediction):} We ran experiments on the VOCA-2012 dataset using the same experimental paradigm as in~\cite{Mathe13}. We used our final SALICON-trained model to predict maps for test images both before and after fine-tuning the model on training images from VOCA-2012. The results summarized in Table~\ref{tab:voca} show that our method, both with and without finetuning, outperforms the 
state-of-the-art~\cite{Mathe13}. This suggests that the task-dependent fixations for this action recognition dataset are highly consistent with free-viewing fixations.

\begin{table}
\footnotesize
\centering
\setlength\tabcolsep{3pt}
\renewcommand{\arraystretch}{1.2}
\begin{tabular}{|c|c|c|}
\hline 
Method & KL & AUC \\
\hline
\hline
HOG detector*~\cite{Mathe13} & 8.54 & 0.736 \\
Judd et al.*~\cite{judd2009learning} & 11.00 & 0.715 \\
Itti \& Koch~\cite{itti2000saliency} & 16.53 & 0.533 \\
central bias~\cite{Mathe13} & 9.59 & 0.780 \\
human~\cite{Mathe13}& 6.14 & 0.922 \\
\hline
PDP(without finetuning) & {\bf 7.92} & 0.845 \\ 
PDP*(with finetuning) & 8.23 & {\bf 0.875} \\ 
\hline 
\end{tabular}
\vspace{2mm}
\caption{VOCA: Performance comparison on KL-divergence and AUC measures. Note that the best performance is achieved by using the fixations of one human observer to predict those of the remaining observers. The results in bold indicate the best-performing methods that do not require human intervention at testing time. (* denotes the methods that have been trained on this particular dataset.)}
\label{tab:voca}
\end{table}
\begin{figure}[ht]
\centering
\begin{tabular}{@{}>{\centering}p{0.2\linewidth}@{}>{\centering}p{0.2\linewidth}@{}>{\centering}p{0.2\linewidth}@{}>{\centering}p{0.2\linewidth}@{}>{\centering}p{0.2\linewidth}}
Image & GT & BMS & SALICON & PDP \\[-1cm]
\vspace{-5cm}
\end{tabular}
\includegraphics[width=\linewidth]{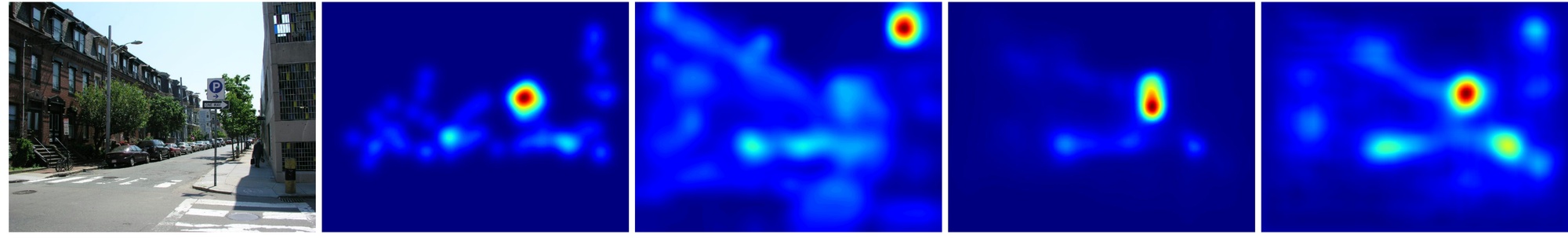} \\
\includegraphics[width=\linewidth]{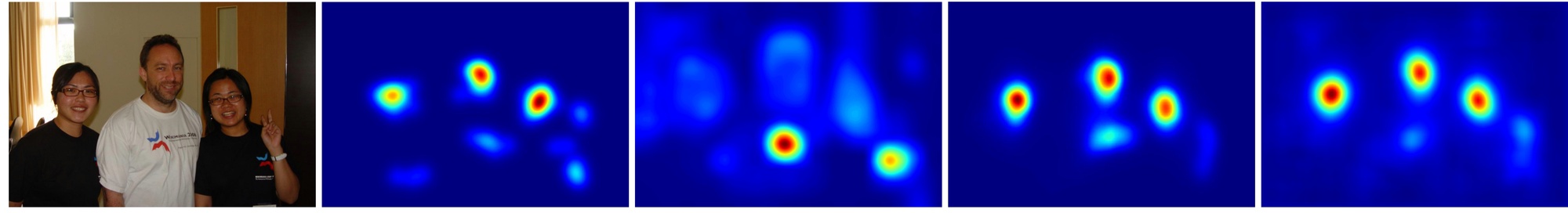} \\
\includegraphics[width=\linewidth]{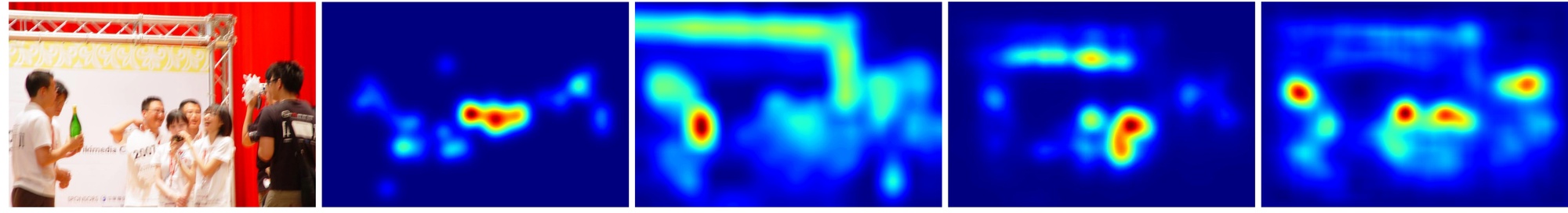} \\
\includegraphics[width=\linewidth]{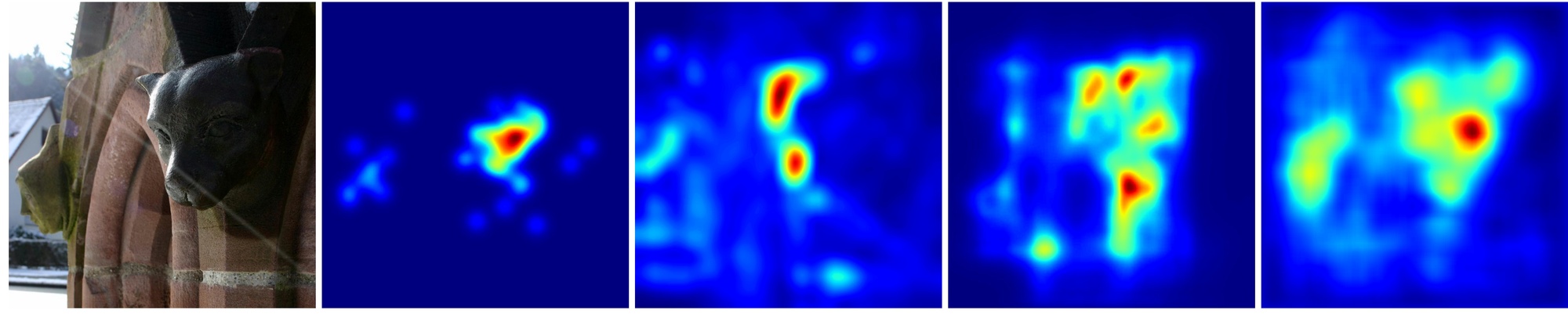} \\
\includegraphics[width=\linewidth]{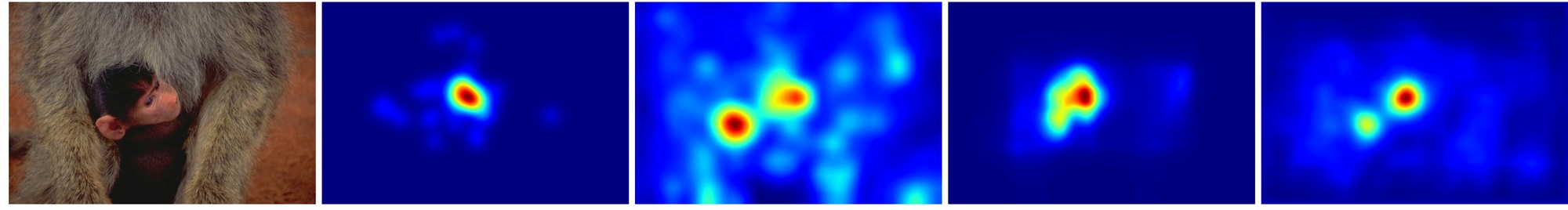} \\
\includegraphics[width=\linewidth]{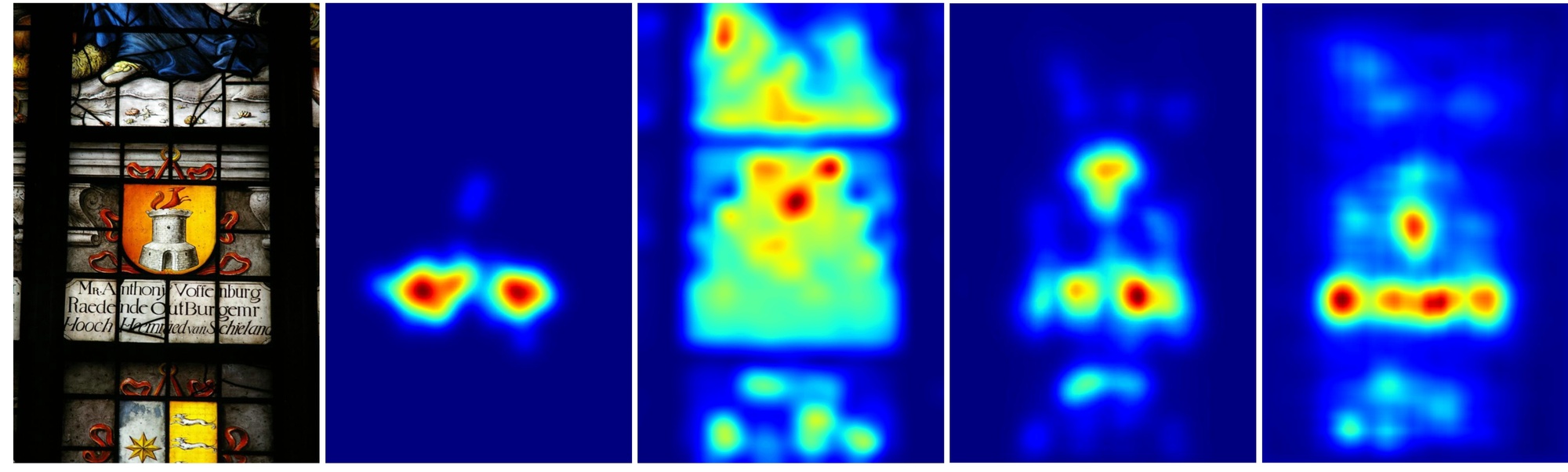} \\
\includegraphics[width=\linewidth]{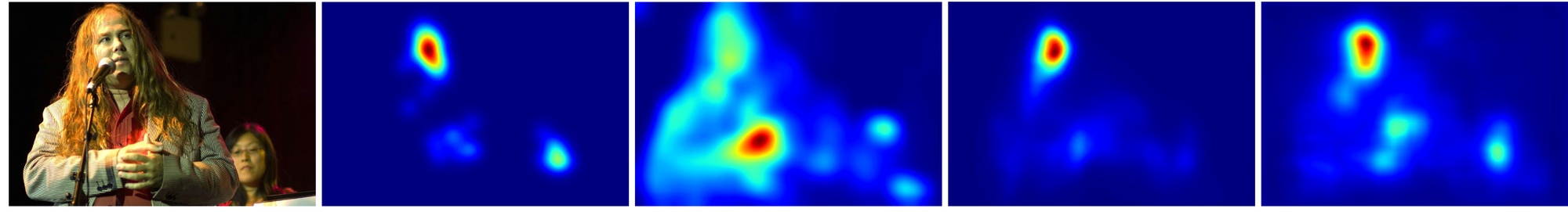} \\
\includegraphics[width=\linewidth]{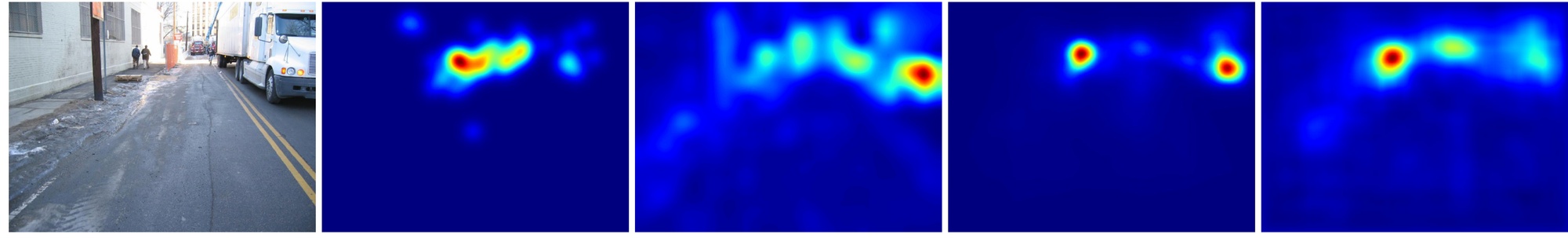} \\
\caption{Comparison of BMS, SALICON, and our proposed PDP method for randomly-sampled images from MIT-1003. GT refers to the ground-truth saliency maps. Note that, to ensure a fair comparison, the PDP results shown here were obtained from a network that was trained only on SALICON images, with no fine-tuning to this dataset.}
\label{fig:qual}
\end{figure}

\subsection{Discussion}
Our probabilistic perspective to saliency estimation is intuitive in two ways. First, attention is competitive as we look at certain regions in the image at the expense of others. Hence, the fixation map normalised over the total visual stimulus can be understood as a spatial probability distribution. Secondly, a probabilistic framework allows the model to account for the noise across subjects and over the data collection paradigm.

To provide qualitative insight, some randomly-chosen predicted maps are shown in Figure~\ref{fig:qual}.
Our method consistently gives high fixation probabilities to areas of high center-surround contrast, and also to high-level cues such as bodies, faces and, to a lesser extent, text. The higher emphasis on bodies and faces as compared to text is likely due to the large number of images containing people and faces in the SALICON dataset.

\begin{figure}[ht]
\centering
\footnotesize
\setlength\tabcolsep{3pt}
\begin{tabular}{m{1cm}m{8cm}}
\pbox{.6\textwidth}{\# of \\samples} & \hspace{0.8cm} Image \hspace{1.5cm} GT \hspace{1.3cm} Prediction \\
$<50$ & \includegraphics[width=0.85\linewidth]{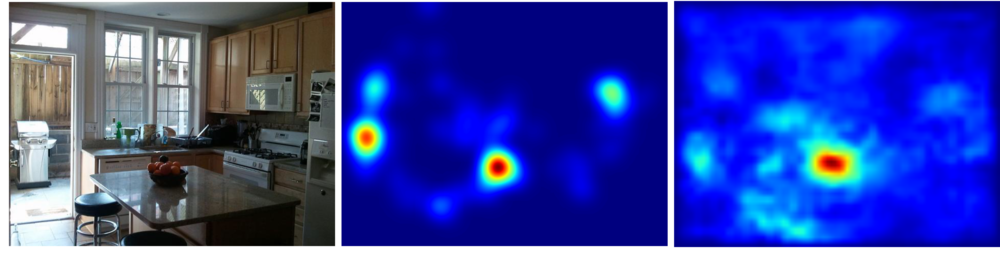} \\
$<50$ & \includegraphics[width=0.85\linewidth]{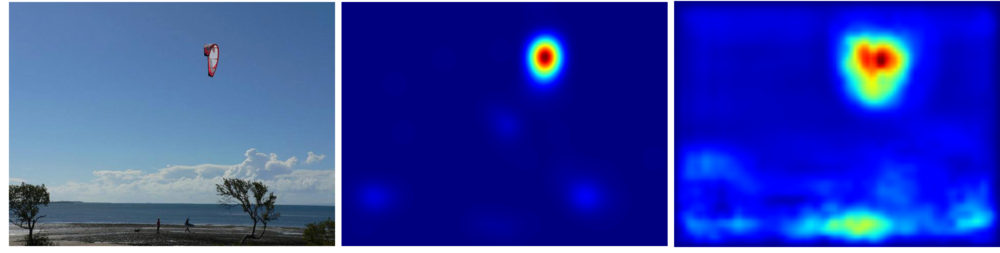} \\
$<100$ & \includegraphics[width=0.85\linewidth]{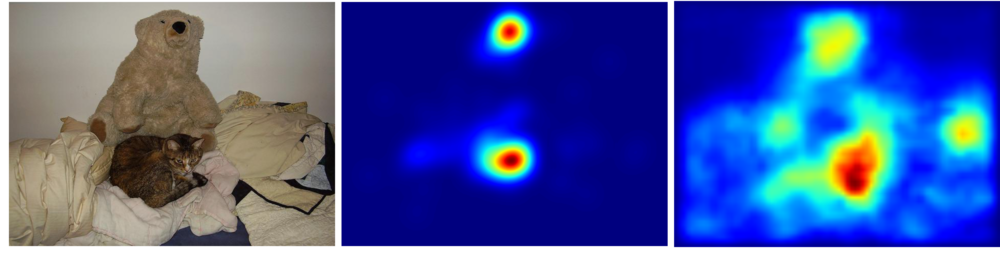} \\
$<100$ & \includegraphics[width=0.85\linewidth]{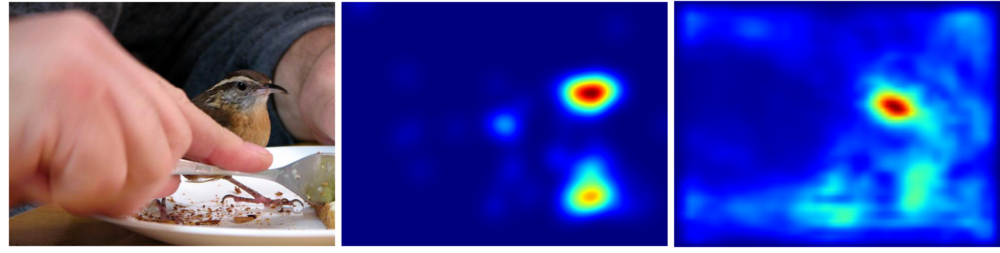} \\
$<400$ & \includegraphics[width=0.85\linewidth]{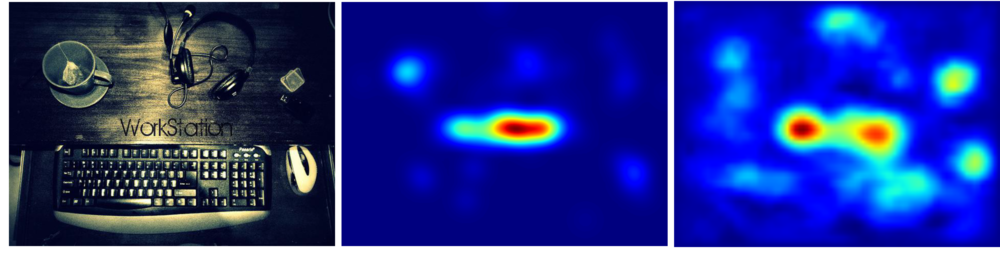} \\
$<400$ & \includegraphics[width=0.85\linewidth]{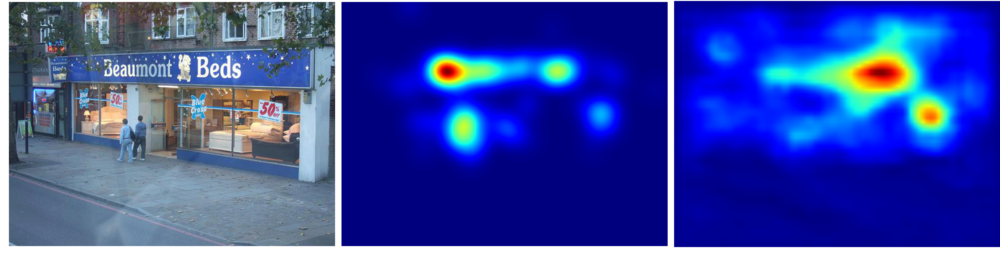} \\
\end{tabular}
\caption{Our method quickly learns that regions of high center-surround contrast, and faces and heads, are salient.}
\label{fig:learning}
\end{figure}

Figure~\ref{fig:learning} shows saliency map predictions for SALICON training images which were obtained on the forward pass after a given number of training images had been used to train the model.
One can see that center-surround contrast cues are learned very quickly, after having seen fewer than 50 images.
Faces (both of animate and non-animate objects) are also learned quickly, having seen fewer than 100 images.
The saliency of text also emerges fairly rapidly. However, the cue is not as strongly identified, likely due to the relatively smaller amount of training data involving text.


\section{Conclusion}\label{sec:conclusion}
We introduce a novel saliency formulation and model for predicting saliency maps given input images.
We train a deep network using an objective function which penalizes the distance between target and predicted maps in the form of probability distributions.
Experiments on four datasets demonstrate the superior performance of our method with respect to other loss functions and other state-of-the-art saliency estimation methods.
They also illustrate the benefit of using suitable learning criteria adapted to this task.

\section*{Acknowledgements}
This work was partly funded by the ERC grant ERC-2012-AdG (321162-HELIOS).


\end{document}